\documentclass[letterpaper, 10 pt, conference]{ieeeconf}  
\usepackage{graphicx}
\usepackage{balance}
\usepackage{cite}
\usepackage{times} 
\usepackage{amsmath} 
\usepackage{amssymb}  
\usepackage{setspace}
\usepackage{caption}
\usepackage{subcaption}
\usepackage{graphics} 
\usepackage{epsfig} 
\usepackage{epstopdf}
\usepackage{booktabs}
\usepackage{multirow}
\AppendGraphicsExtensions{.tif}
\IEEEoverridecommandlockouts 
\title{\LARGE \bf Design of a Robust Stair Climbing Compliant Modular Robot to Tackle Overhang on Stairs}

\author{Ajinkya Bhole$^{1}$,  Sri Harsha Turlapati$^{1}$, Rajashekhar V. S $^{1}$, Jay Dixit$^{1}$, Suril V. Shah$^{2}$, Madhava Krishna K$^{1}$ \thanks{Ajinkya Bhole, Sri Harsha Turlapati, Rajashekhar V. S, Jay Dixit and K Madhava Krishna are with the Robotics Research Lab, IIIT-Hyderabad, TS 500032, India and Suril V. Shah is with Department of Mechanical Engineering, IIT Jodhpur, India. Corresponding Author: {surilshah@iitj.ac.in}}
}

\begin{document}

\maketitle

\begin{abstract}
This paper discusses the concept and parameter design of a Robust Stair Climbing Compliant Modular Robot, capable of tackling stairs with overhangs. Modifying the geometry of the periphery of the wheels of our robot helps in tackling overhangs. Along with establishing a concept design, robust design parameters are set to minimize performance variation.\\
The Grey-based Taguchi Method is adopted for providing an optimal setting for the design parameters of the robot. The robot prototype is shown to have successfully scaled stairs of varying dimensions, with overhang, thus corroborating the analysis performed. 
\end{abstract}

\section{Introduction}
\begin{figure}[t]
\centering
\includegraphics[height=0.45\linewidth]{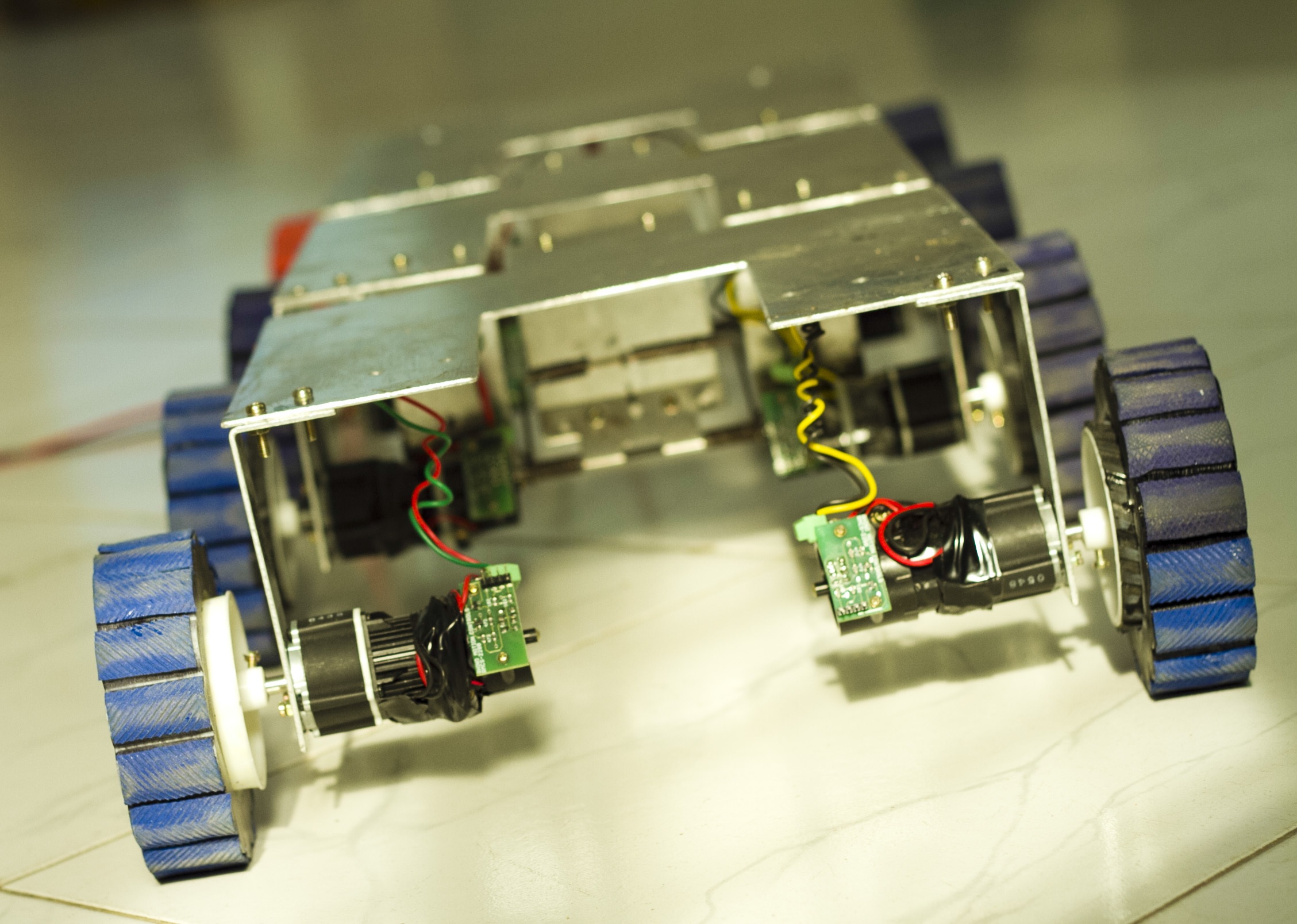}
\caption{Stair Climbing Compliant Modular Robot Prototype}
\label{Robot_1}
\vspace{-8mm}
\end{figure}

Stair traversal is a critical requirement for search and rescue robots as many of the terrestrial mission scenarios occur in urban settings with stairs. In this paper, we present the concept design of a  Stair Climbing Compliant Modular Robot (Figure \ref{Robot_1}), that can tackle overhangs on stairs and also equip it with robust design parameters to make its performance insusceptible to varying step dimensions.

Although legged mechanisms have shown tremendous advantage in traversing uneven terrain and climbing over obstacles, these robots have complex designs, need complex control strategies and are sluggish. A wheeled mechanism is a better option due to its design simplicity, lower power consumption and quick mobility. This justifies our choice of a wheeled robot.

Power consumption in an urban search and rescue robot is a cardinal performance factor, as the robot must not stop in the middle of a mission by draining its power source. Hence, it is essential to develop energetically efficient robots for such operations. The wheel diameter is one of the major design factors which affects power consumption. Large wheels often become bulky, require high driving torque and thus result in greater power consumption than smaller wheel. In the following development, we call the robots which use wheels with diameter smaller than the stair riser height as `Class I' robots and, `Class II' robots the vice-versa. 

Our previous works \cite{avi,harsha}, the Rocker-Bogie \cite{rocker}, Shrimp \cite{shrimp} and Octopus \cite{octopus} are examples of Class I robots while robots like Asguard\cite{asguard}, Genbu \cite{genbu} , Rhex \cite{rhex}, Impass \cite{impasss}, Loper \cite{loperr} are Class II robots. Class I robots have the advantage of low power consumption over Class II robots, but one drawback with Class I robots is that these mechanisms can get jammed under a step overhang (explicated in Sec. II). Overhangs/Nosings are protrusions beyond the face of the stair riser, on the front edge of a stair. Class II robots owing to their larger wheels do not face this problem as they are able to make direct contact with the vertical face of the overhang or the next step tread. This drawback of Class I robots, to the best of our knowledge has not been identified in any previous works on stair climbing robots. In this work we undertook the problem of overcoming overhangs by keeping intact the feature of Class I robots, i.e. wheel diameter lesser than the stair riser height (in fact lesser than three-fourths of the stair riser height), thus maintaining the low power consumption advantage. This endeavor was achieved by modifying the design of the wheels of our robot, by blending equally spaced circular arcs on the periphery of the wheel.

As mentioned earlier, we also dressed our robot with robust design parameters so that its performance is least affected due to the varying step dimensions. For a robust design of the robot, we define the desired performance metrics which depend on certain critical controllable and uncontrollable design parameters. We consider the overall power consumed by the robot as a performance metric. This depends on controllable design parameters, some of which are length of the module, radius of the wheel and number of arcs on the wheel and among uncontrollable design parameters are the varying stair dimensions and the coefficient of friction. The wheel design modification of integrating arcs on the periphery of the wheel introduces transverse alterations in the path of the robot. Thus, along with the overall power consumption, we also set the amplitude and frequency of the transverse alterations as the performance metrics. Altogether, we set the controllable design parameters such that the power consumption and its variation, and also the amplitude and frequency of transverse alterations arising from the wheel design are reduced. For this, the Grey-based Taguchi Method \cite{senthi, grey2, grey3, madm}, a Multi-Objective Optimization method is used, which takes care of the robustification as well as combining multiple
performance attributes together for the decision making process, to obtain an optimal set of controllable design parameters.
\begin{figure*}[t]
\centering
\includegraphics[width=\linewidth]{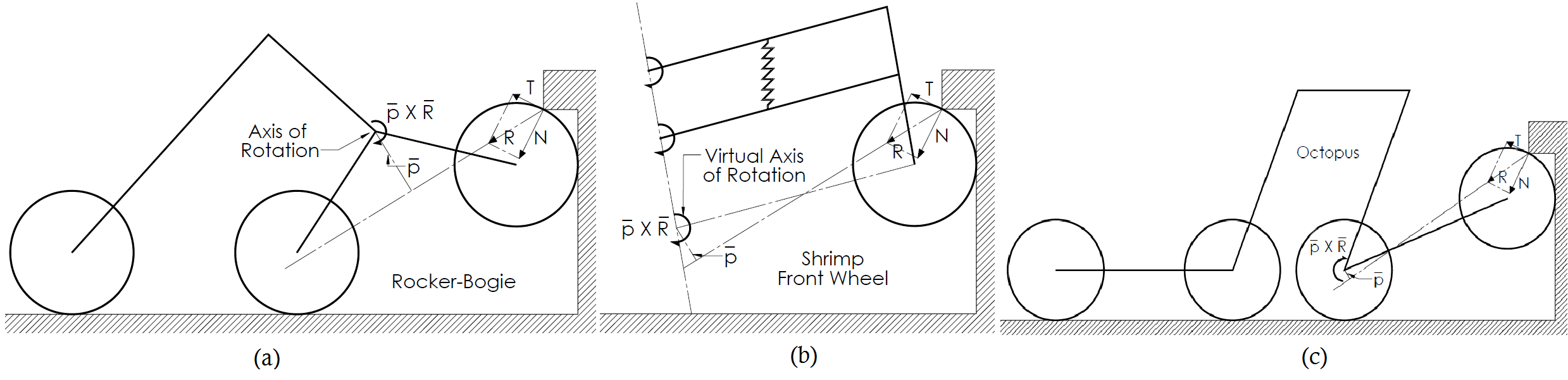}
\caption{Some eminent Class I robots which could get jammed under stair overhang}
\label{Fail}
\vspace{-4mm}
\end{figure*}

A body of work has been devoted to the concept design of robots for stair climbing, but none of them to the best of our knowledge, discussed specifically on the challenge of tackling overhangs, with a rigour like ours, thus providing a novelty to our work. Also, many previous works on stair climbing robots have considered random stair dimensions or dimensions which suit their robot designs, for their analysis. We have considered stair dimensions which follow the International Building Code (IBC) \cite{ibc} and thus our work has a high impact for rescue operations in urban settings. 

The paper contributes as follows. Firstly, a solution to overcome overhang is proposed by modifying the wheel design. Secondly, it poses the parametric design of the robot as a multi-objective optimization problem. Since a functional characterisation of the same is almost intractable, it makes use of statistical methods \cite{senthi, grey2, grey3, madm} to solve for the optimal design. Setting the range for the design factors for this optimization problem forms a cardinal part of this work. This is a critical job, as setting an arbitrary range can lead to some configurations in which the robot can get stuck or even not climb. With a careful analysis of all the design factor we have logically backed every design factor's range selection. As a consequence of this design procedure, a prototype was fabricated and is shown to climb stairs and overhangs of varying dimensions many of them long and protruding. To the best of our knowledge, such a formal method of posing wheel design to tackle overhangs as a statistical optimization formulation and the concomitant analysis and detailed experimental verification does not appear in the surveyed literature.

The paper is organised as follows: Section \ref{conceptdesign} discusses a concept design for tackling overhangs on stairs. A model for the parametric design of the robot is setup in Section \ref{pd}. Section \ref{doe} presents design of experiments for evaluating performance metrics through Taguchi's Orthogonal Array Experiments. Section \ref{exp} confirms the analysis performed by practical experimentation. Finally, conclusions and scope of future work are discussed in the Section \ref{future}.

\section{Concept Design} \label{conceptdesign}
The previous design of our stair climbing compliant modular robot \cite{harsha} with wheels of diameter lesser than the stair riser height got jammed under the overhangs.
\begin{figure}[h]
\vspace{-1mm}
\centering
\includegraphics[width=\linewidth]{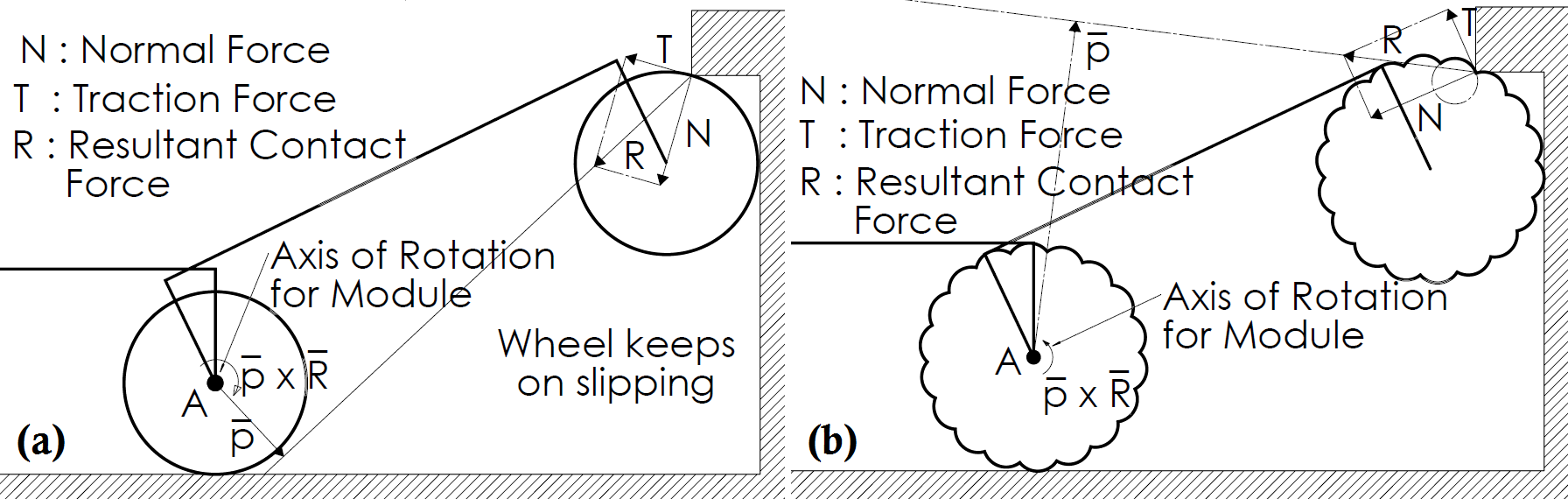}
\caption{(a) Circular wheel failing to provide the desired counter-clockwise moment about axis of rotation for Module (b) Modified wheel design providing the desired counter-clockwise moment}
\label{jamsol}
\vspace{-6mm}
\end{figure}

As can be seen in Figure \ref{jamsol} (a), the first module gets jammed under the overhang as the external contact force $R$ obtained from the ground on the wheel fails to provide a counter-clockwise moment about the axis of rotation of the module i.e. axis A. This prevents the module from folding inwards thus disabling climbing. As can be seen in Figure \ref{Fail}, the same problem would be faced by some of the eminent Class I robots.

A counter-clockwise moment can be easily achieved by changing some of the design parameters of the vehicle. For example, decreasing the length of the module or increasing the radius of the wheel to such an extent that the contact force starts providing the desired counter-clockwise moment are viable options. But as discussed in the forthcoming Section III.C.b., in our case, the module length is constrained by a certain lower bound, exceeding which the module tips over while climbing. Increasing the radius of the wheel also is an unfavourable option, as this makes the robot quite bulky, increasing the power consumed. Similar design constraints can also occur in other Class I robots.

The exercise of integrating knowledge of the geometry of stairs and mechanical interaction into a solution to the stair climbing problem usually comes in one of two flavors: we either use design (eg. \cite{harsha, rocker, shrimp}) or generate plans (e.g., Rhex \cite{rhex} uses a special algorithm). Shape is an easily alterable design freedom with potential benefits both in terms of simplicity and robustness, hence we explored this. We modified the wheel design of our robot. As can be seen in Figure \ref{jamsol} (b), blending arcs on the circumference of the wheel changes the direction of the resultant contact force obtained from the overhang, providing a counter-clockwise moment about the axis of rotation of the folding module.
\begin{figure}[h]
\centering
\begin{minipage}[b]{0.33\linewidth}
\includegraphics[width=\linewidth,trim = 64mm 130mm 245mm 50mm, clip]{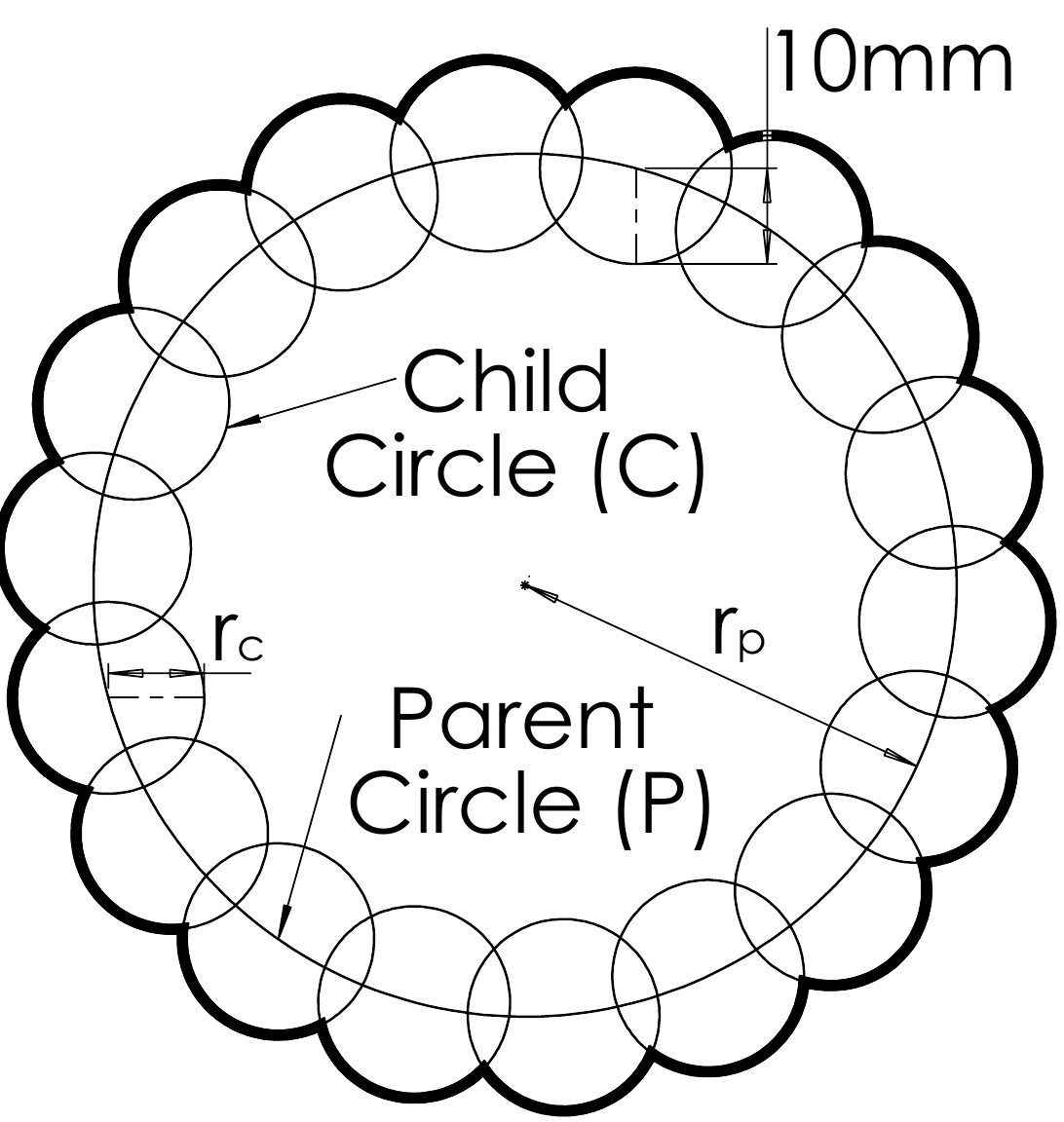} 
\caption{Wheel Design}
\label{Wheel}
\end{minipage}
\begin{minipage}[b]{0.65\linewidth}
\includegraphics[width=\linewidth,trim = 63mm 80mm 58mm 40mm, clip]{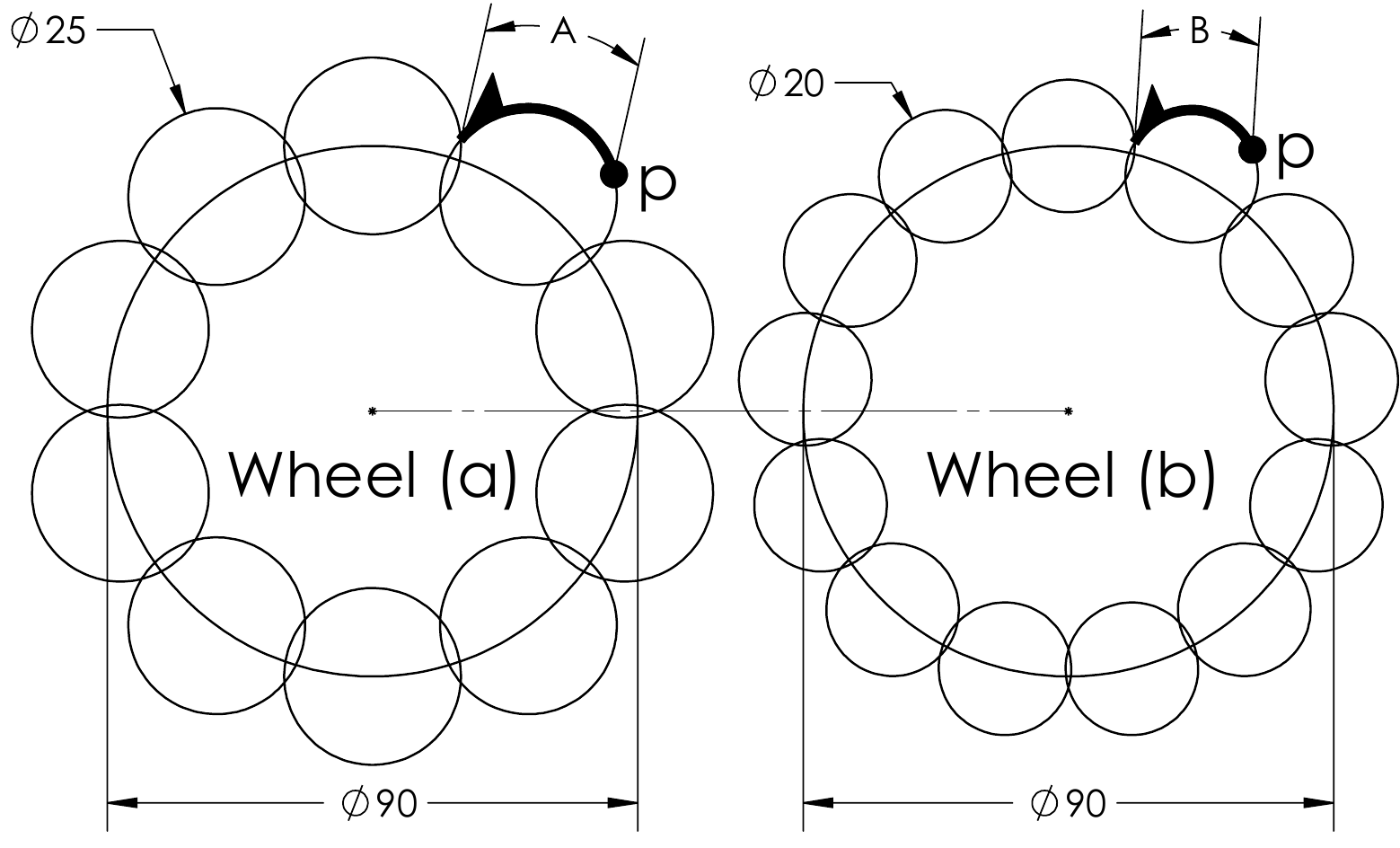} 
\caption{Child Circle size selection}
\label{compare}
\end{minipage}
\vspace{-10mm}
\end{figure}

As shown in the Figure \ref{Wheel}, the wheel is designed by drawing a parent circle $P$ of radius $r_p$ and evenly spaced child circles $C$ of radius $r_c$ on the circumference of the parent circle. Varying the radius of parent circle and the number of child circles changes the shape of the circumference of the wheel. We next describe the child circle size selection.

Shown in Figure \ref{compare} are two wheel designs with different radii for child circle (Wheel (a) with a greater radius). The radius of the parent circle was kept same for both the wheels and the number of child circles was set to a minimum value so as to not allow the child circles to be disjoint (this condition is explicated in Section III.C.c) If for both the wheels, a point $p$ on the child circle contacts the lower tip of the overhang where slippage starts occurring due to the inability of the contact force to produce counterclockwise moment about axis of rotation of module, then, Wheel (a) will take a greater time to come out of this slippage. This is because the Wheel (a) needs to cover a greater arc length (A $>$ B) before coming to a point on the wheel periphery which ends slippage. Hence, it is better to keep the radius of the child circle as low as possible. 

However, for practical purpose this radius cannot be set too low. This is because wheels usually require a rubber covering which helps to provide traction. Setting child circle radius too low will result in the rubber cover, mainly forming the child circle shape. Rubber is deformable and would not result in providing the desired direction of the contact force, thus inapt to provide the essential counterclockwise moment. Such geometry would in fact become almost circular and thus resemble a normal circular wheel which fails in overcoming overhang. Thus we set a reasonable diameter for the child circle, one which is comparable with the dimension of the overhang and fixed its value (to 20mm) to reduce the number of variables which result in the changing wheel geometry and thus lower its design complexity.

Eminent robots like Asguard \cite{asguard} and  Whegs \cite{whegscock} also have adopted the strategy of change in the wheel design to tackle stairs. As opposed to their design, we have tried to keep the wheel as close to a circle as possible. This provides greater amount of contact with the ground, thus better traction, speed and low power consumption. In order to demonstrate the power efficacy, we fitted our robot with the wheel designs used in \cite{asguard} and \cite{whegscock}, keeping the same outer diameter. We ran simulation experiments and measured the power consumption for a run with constant motor speed of 10 $rpm$ on a flat terrain for 5 $sec$. The results depicted in Table \ref{wheelcomptable}, clearly shows the superiority of our proposed  wheel design in comparison to the ones proposed in \cite{asguard} and \cite{whegscock}. Along with the low power consumption, robot with our wheel design also traversed a longer distance due to its greater circularity, thus showcasing its efficacy. Circular arc geometry, instead of spiky geometry, on the periphery of the wheel ensures a gradual variation in the direction of the contact force from the ground. This, consequently, avoids undesired fluctuations in the torque required from the motor. \vspace{-1mm}
\begin{table}[h]
\centering
\caption{Comparion with other eminent wheel designs}
\label{wheelcomptable}
\resizebox{\linewidth}{!}{%
\renewcommand{\arraystretch}{1.5}
\begin{tabular}{cccc}
\hline
& Our Wheel & Asguard Type Wheel & Whegs Type Wheel \\ \hline
\begin{tabular}[c]{@{}c@{}}Power Consumption \\ (Watt)\end{tabular} & 0.1544    & 0.3252           & 0.8516 \\ \hline       
\end{tabular}
}
\vspace{-3mm}
\end{table}
\section{Parameter Design} \label{pd}
The multiple performance goals of increasing the robustness of the robot and reducing the transverse alterations which result from the wheel design can be achieved by setting up a proper model for the parametric design. Posing this as an optimization problem to solve for the optimal parameters is quite difficult owing it to the complexity in the design.   Statistically evaluating the performance based on different experiments intuitively points in the direction of the optimal design choices. We achieve this by using the Grey-based Taguchi Method, which is a widely adopted statistical method for solving Multiple Attribute Decision Making (MADM) problems \cite{madm,senthi,grey2,grey3}. This method is a combination of the Taguchi Method \cite{taguchi} and the Grey Relational Analysis (GRA) \cite{deng}. The Taguchi method takes care of providing a robust design for the robot and the GRA helps combining various performance metrics into a single metric for decision making.

The Grey-based Taguchi Method involves defining the desired performance metrics (response) and the various controllable and uncontrollable (noise) factors and their levels in the system, influencing the defined performance metrics. These performance metrics and the design factors are discussed in the immediate subsections.
\subsection{Performance Attributes}
\begin{figure}[h]
\vspace{-3mm}
\centering
\includegraphics[height=25mm,trim = 180mm 195mm 110mm 135mm, clip]{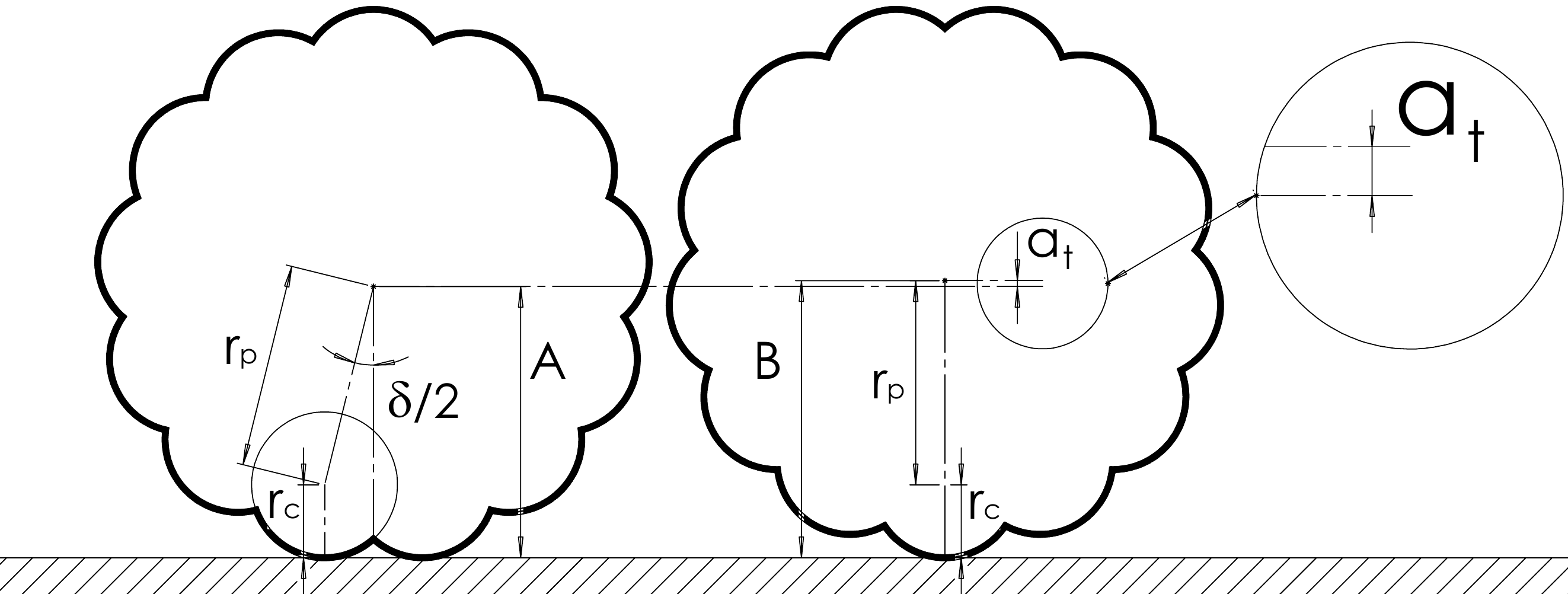} 
\caption{Geometry for calculating Amplitude of Transverse Alterations}
\label{amp}
\vspace{-4mm}
\end{figure}

The power consumed by the 8 motors of the robot in climbing 4 steps ($p_r$), the amplitude ($a_t$) and frequency of transverse alterations ($f_t$) are considered as the performance attributes. From Figure \ref{amp}, $a_t$ is given by
\begin{equation}\label{Eq:Amplitude}
\centering
a_t = B-A = r_p (1-cos(\delta /2)) \nonumber
\end{equation}

The frequency of transverse alterations per rotation of the wheel is equal to the number of child circles on the wheel.
\begin{equation}
\centering
f_t = n_c \nonumber
\end{equation}

\subsection{Noise Factors}
\begin{figure}[b]
\centering
\resizebox{\linewidth}{!}{
  \begin{tabular}{ccc}
    \includegraphics[height=0.8\linewidth,trim = 185mm 100mm 265mm 190mm, clip]{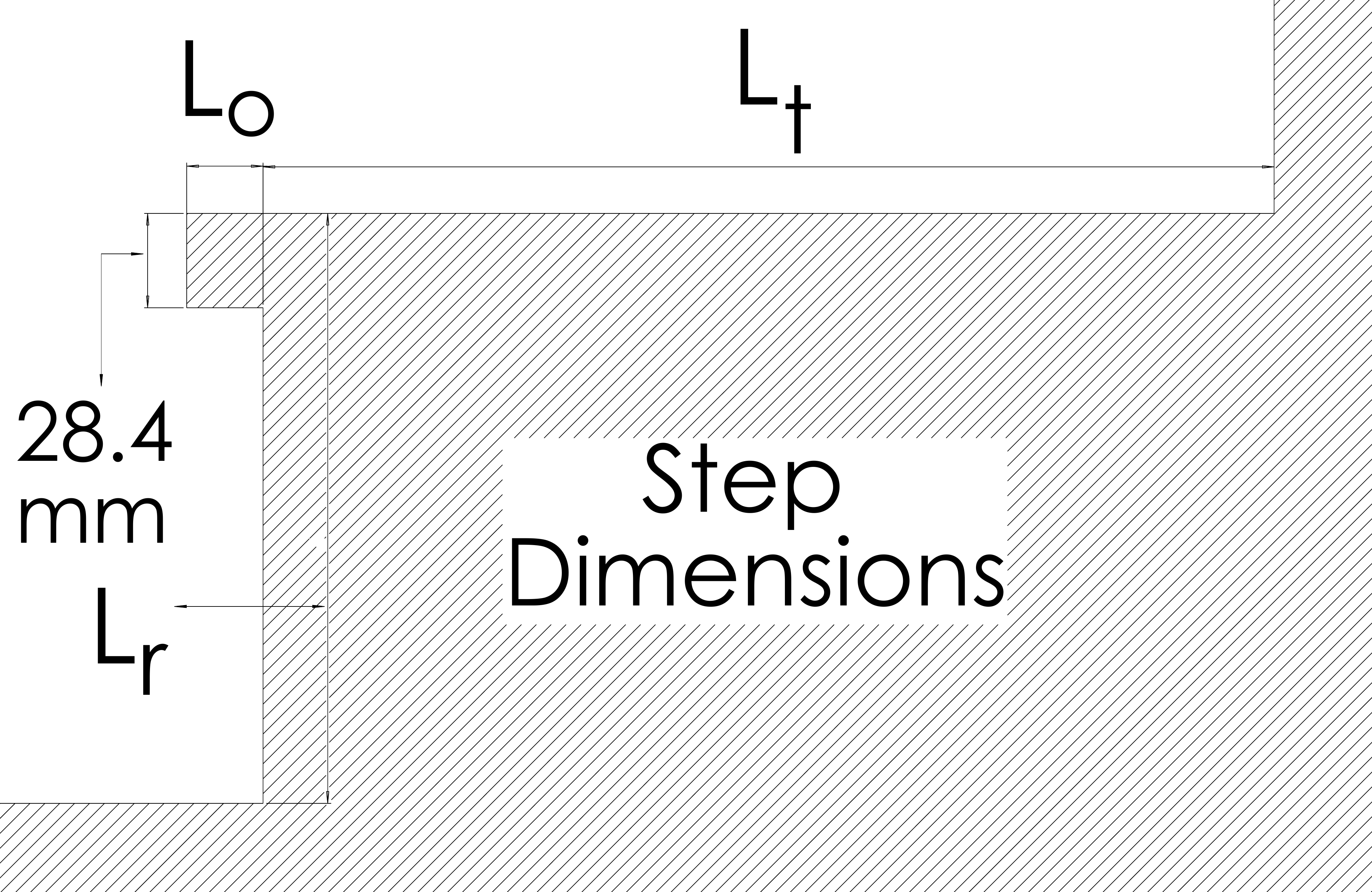} &
    \includegraphics[height=0.6\linewidth,trim = 80mm 80mm 60mm 70mm, clip]{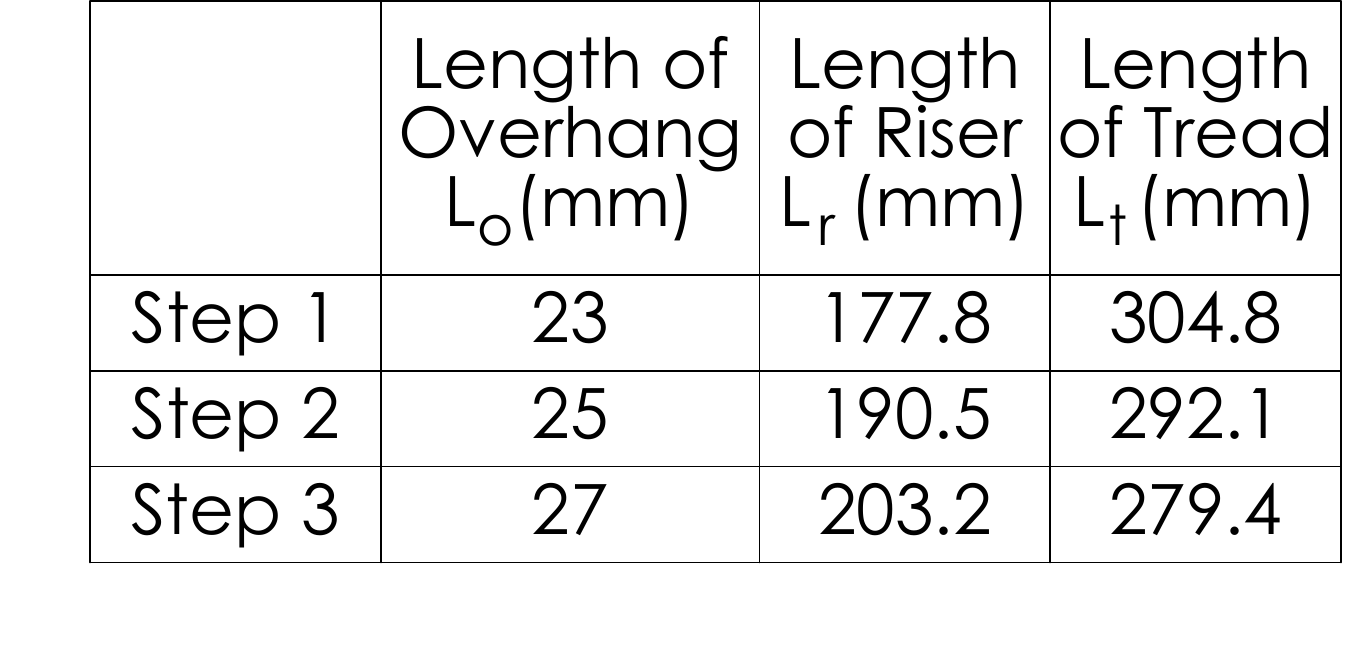} & \\
  \end{tabular}
  }
  \caption{Varying Step Dimensions}
  \label{Noise}
\end{figure}

The factors affecting the performance of a system which are beyond the control of the designer are termed as noise factors. The Grey-based Taguchi Method provides a design setting, making the performance of the system least sensitive to these factors. The coefficient of friction and the varying stair dimensions are the noise factors in our system. We consider three different stair dimensions shown in the Figure \ref{Noise} which follow the IBC. For the simplicity of analysis, we assumed the coefficient of friction to be constant.
\subsection{Control Factors}
\begin{figure}[h]
\vspace{1mm}
  \begin{subfigure}[b]{0.49\linewidth}
  \centering
    \includegraphics[height=20mm,trim = 130mm 310mm 340mm 40mm, clip]{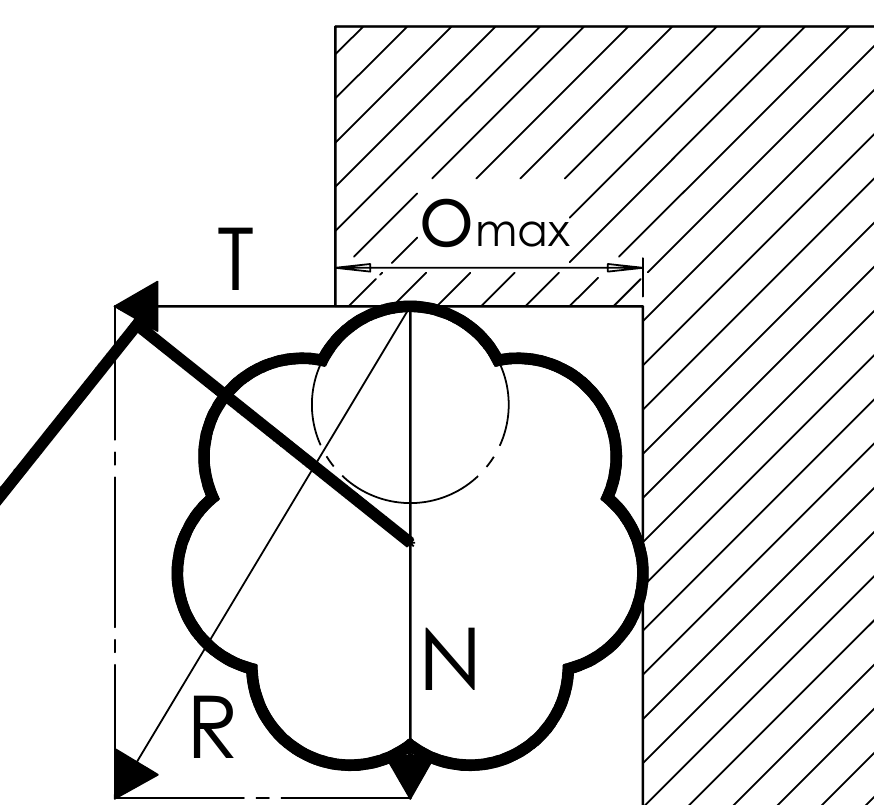}
    \caption{Normal Force acting vertically downwards\\}
    \label{Rmin_1}
  \end{subfigure}
  \begin{subfigure}[b]{0.49\linewidth}
  \centering
    \includegraphics[height=20mm,trim = 100mm 220mm 300mm 90mm, clip]{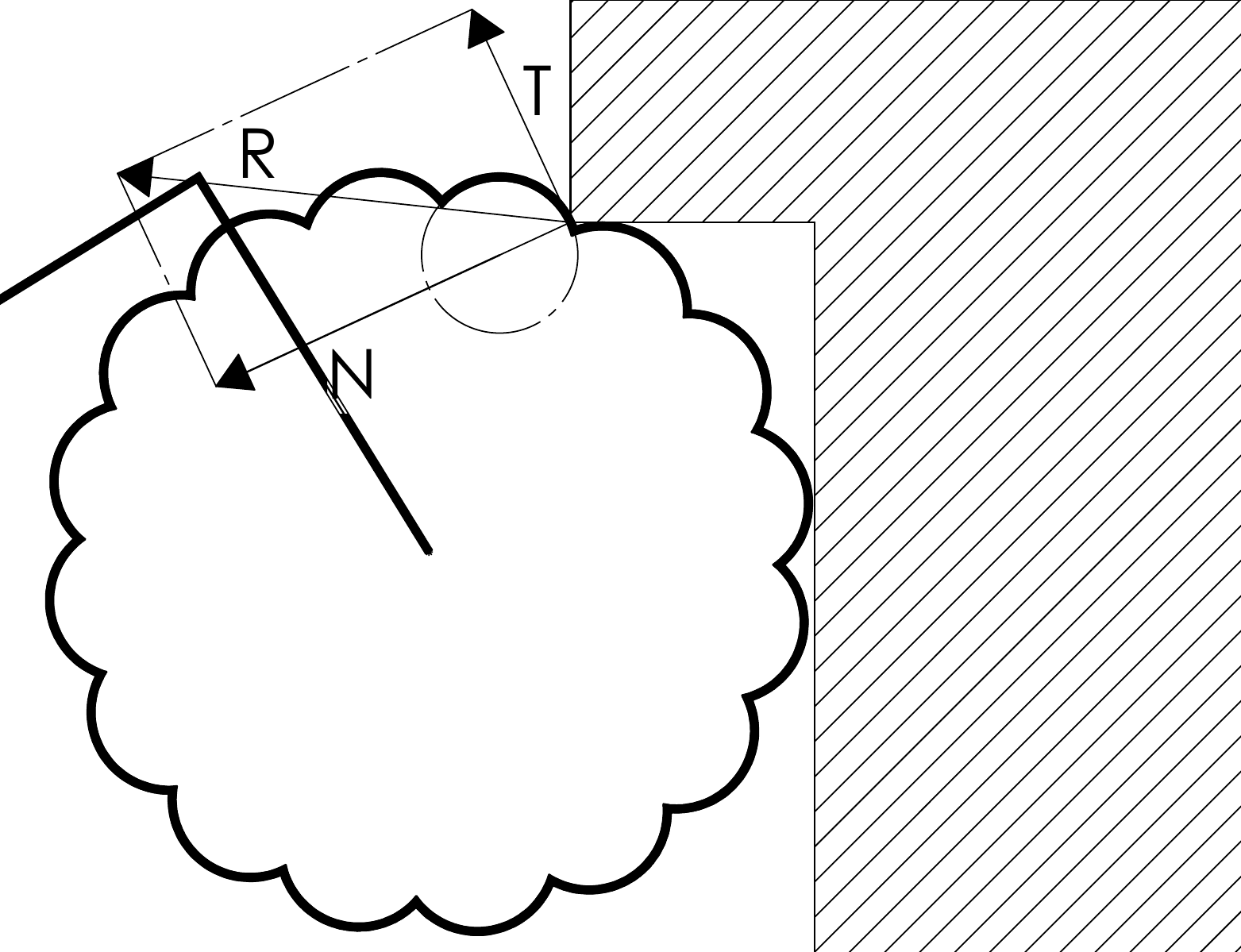}
    \caption{Normal Force shifted clockwise towards the axis of rotation of module}
    \label{Rmin_2}
  \end{subfigure}
  \caption{Concept for setting the lower bound on the radius of Parent Circle}
  \label{Rmin}
  \vspace{-6mm}
\end{figure}

Control factors are the design parameters which can be practically controlled. It is necessary to identify the factors which influence the performance of the system and find the range in which they lie. The various control factors and their ranges are described ahead. 
\paragraph{Parent Radius of Wheel ($r_p$)}
Small-sized wheels create problems in geometrical trafficability of the robot. As Figure \ref{Rmin} illustrates, in case of a wheel radius lesser than the overhang dimension, the normal reaction force obtained from the overhang acts vertically downwards. Keeping the radius greater than the overhang dimension, shifts the normal force from a vertically downward direction, towards the axis of rotation of module, making it easier for the resultant contact force to provide a counter-clockwise moment about the axis of rotation of module. Therefore, the lower bound, of the overall radius of the wheel was set as the maximum overhang length allowed according to the IBC Standards. 

Many of the stair climbing vehicles usually make use of large wheels with diameter greater than the step riser dimensions \cite{asguard,genbu,rhex,impasss,loperr}. This makes the vehicle quite bulky, consuming a lot of power. Considering this, we undertook the challenge of stair climbing and tackling overhangs, with wheels smaller than step riser and set the upper bound on the overall wheel diameter, as three quarters of the minimum riser dimension considered in our analysis. This gave us the following range for the levels for the parent radius:
\begin{equation}\label{Eq:R_lb}
\centering
r_p + r_c \geq o_{max}  \nonumber
\end{equation}
\begin{equation}\label{Eq:R_up}
\centering
2(r_p + r_c) \leq \frac{3}{4}(riser_{min}) \nonumber
\end{equation}
\begin{equation}\label{Eq:R_bound}
\centering
21.25\ mm \leq r_p \leq 56.67\ mm 
\end{equation}
here, $o_{max} = 31.75\ mm$ is the maximum overhang dimension allowed according to the IBC and $riser_{min} = 177.8\ mm$ is the minimum step riser dimension considered in our analysis.

\begin{figure}[h]
\centering
\resizebox{\linewidth}{!}{
  \begin{tabular}{ccc}
    \includegraphics[width=0.8\linewidth,trim = 355mm 109mm 65mm 110mm, clip]{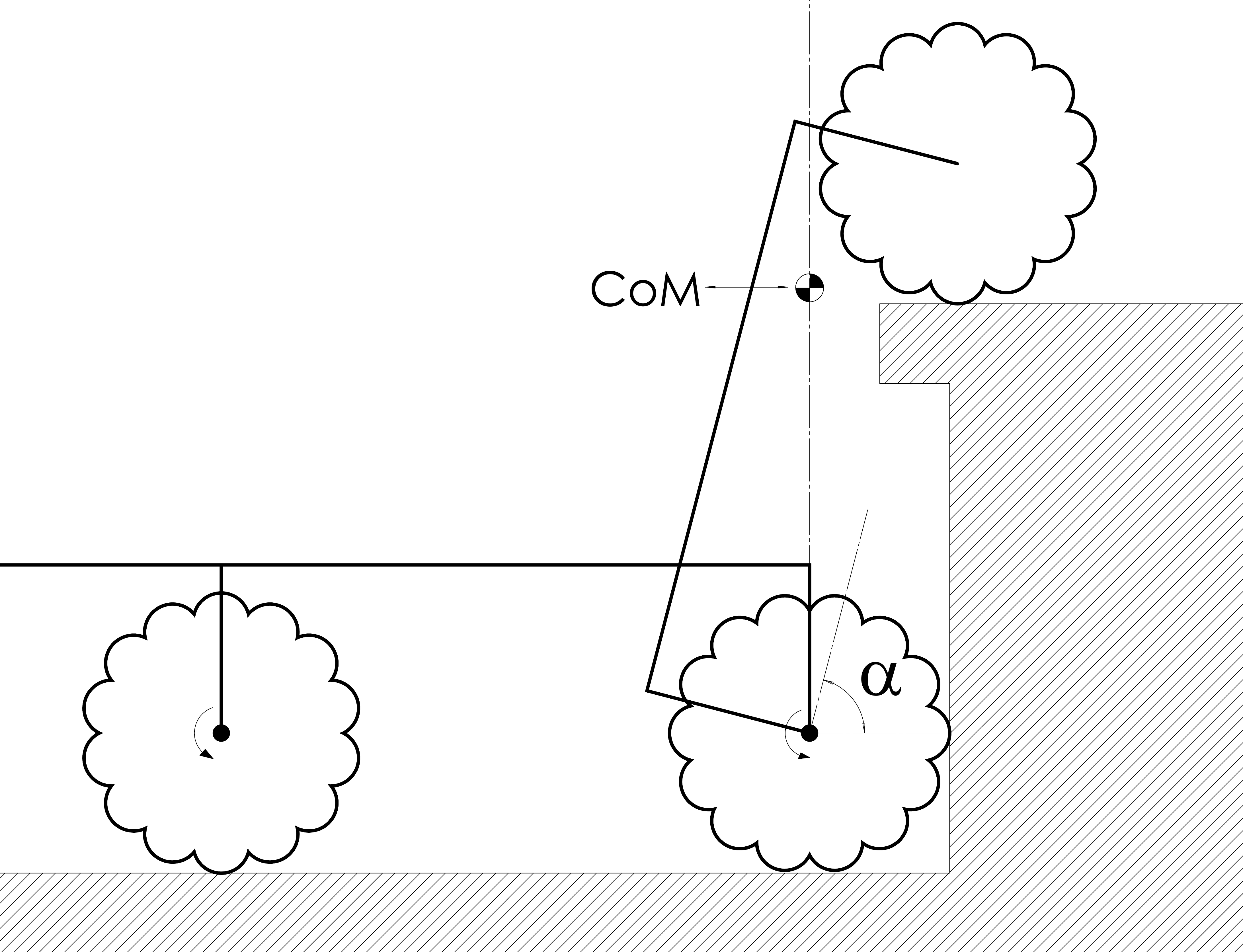} &
    \includegraphics[width=0.8\linewidth,trim = 355mm 100mm 65mm 110mm, clip]{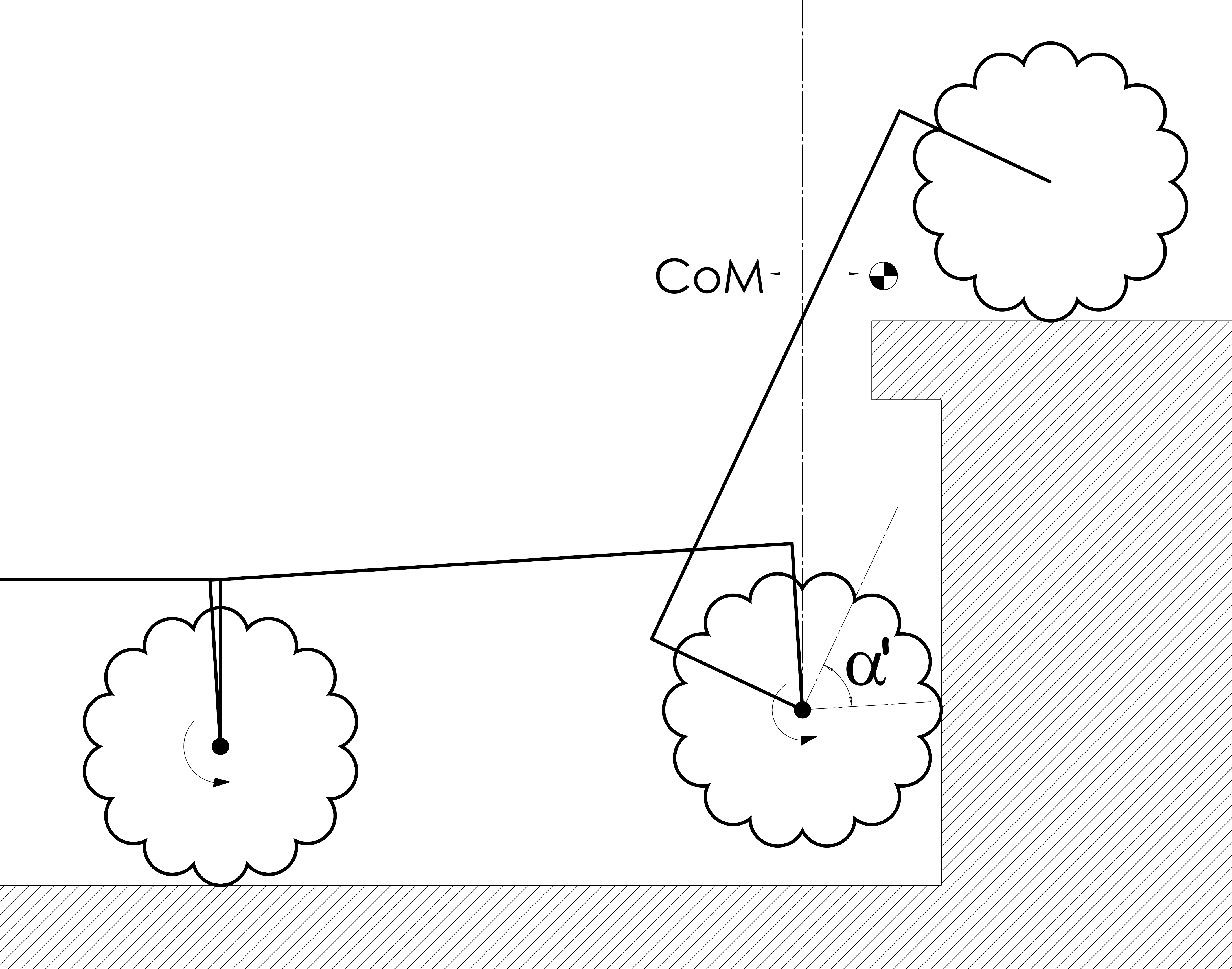} & \\
  \end{tabular}
  }
  \caption{Concept for setting a lower bound on the module length}
  \label{Lmin}
  \vspace{-4mm}
\end{figure}

\paragraph{Length of Module ($l_m$)}
As was discussed in \cite{avi}, the first module climbs a step till it crosses a limiting angle, in which case the module tips over. This occurs because the center of mass of the module climbing up the stair, starts generating a counter-clockwise moment about the axis of rotation of the module. As can be seen in Figure \ref{Lmin}, keeping the length of the module in such a way that the center of mass of the module climbing generates clockwise moment about its axis of rotation until its wheels overcome the overhang and contact the stair tread, assuredly avoids tip over as after this, the relative angle $\alpha$ between the first and the second module starts decreasing subsequently. The lower limit for the length of the module was set considering this requirement.

For a certain range of the module length, configurations similar to the one shown in Figure \ref{Lup} arise. Such a configuration is undesirable because there are no contact forces pushing the vehicle forward on the stairs. For this reason, at least one set of the wheels must be present on the step tread, so that they are able to push/pull the other wheels on the step riser. This can be achieved by setting the length lesser than the one shown in Figure \ref{Lup}. Following range was obtained for the module length considering these requirements:
\begin{equation}\label{Eq:Len_bound}
\centering
210.96\ mm \leq l_m \leq 352.37\ mm 
\vspace{-2mm}
\end{equation}

\begin{figure}[t]
\centering
\includegraphics[width=\linewidth]{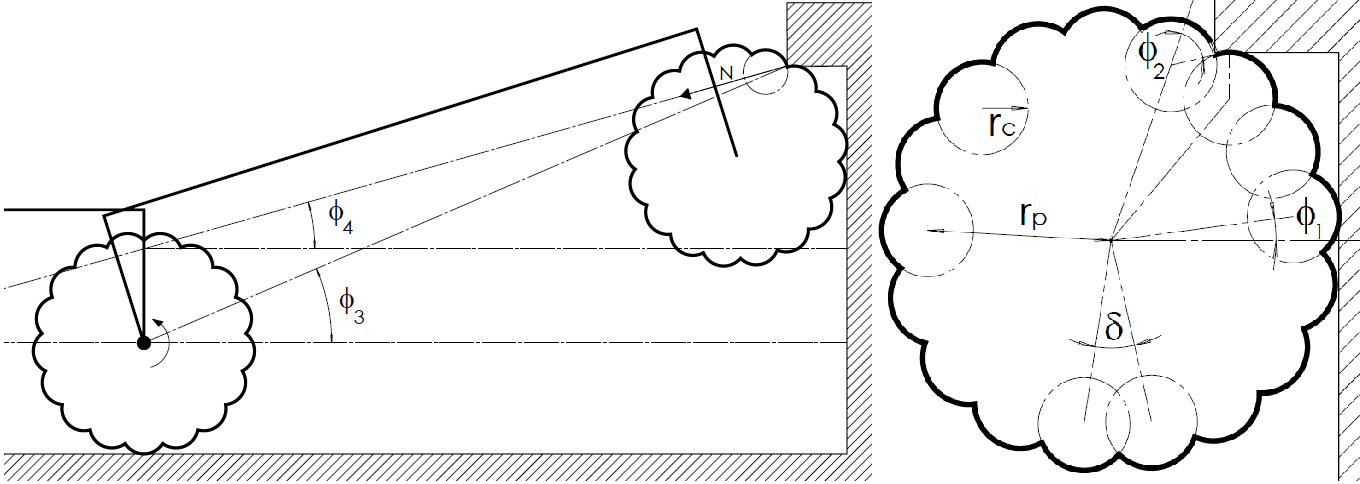} 
\caption{Geometry for setting an upper bound on the number of child circles}
\vspace{-5mm}
\label{Nup2}
\end{figure}

\paragraph{Number of Child Circles ($n_c$)}
As illustrated in Figure \ref{N_lb}, the lower bound $n_{lb}$ for the number of child circles was chosen such that the parent circle does not occupy any part of the circumference of the wheel i.e. by not allowing the child circle to be disjoint. This is because wheel slip occurs on the sections on the circumference occupied by the parent circle, while on the overhang. This measure does not completely cancel the slippage but reduces it. Reducing slip is important since having repeated slipping as the robot climbs stairs is going to cause more energy consumption. The condition for lower bound can be given as follows:
\begin{figure}[h]
\centering
\begin{minipage}[b]{0.45\linewidth}
\includegraphics[width=\linewidth ,trim = 80mm 90mm 115mm 70mm, clip]{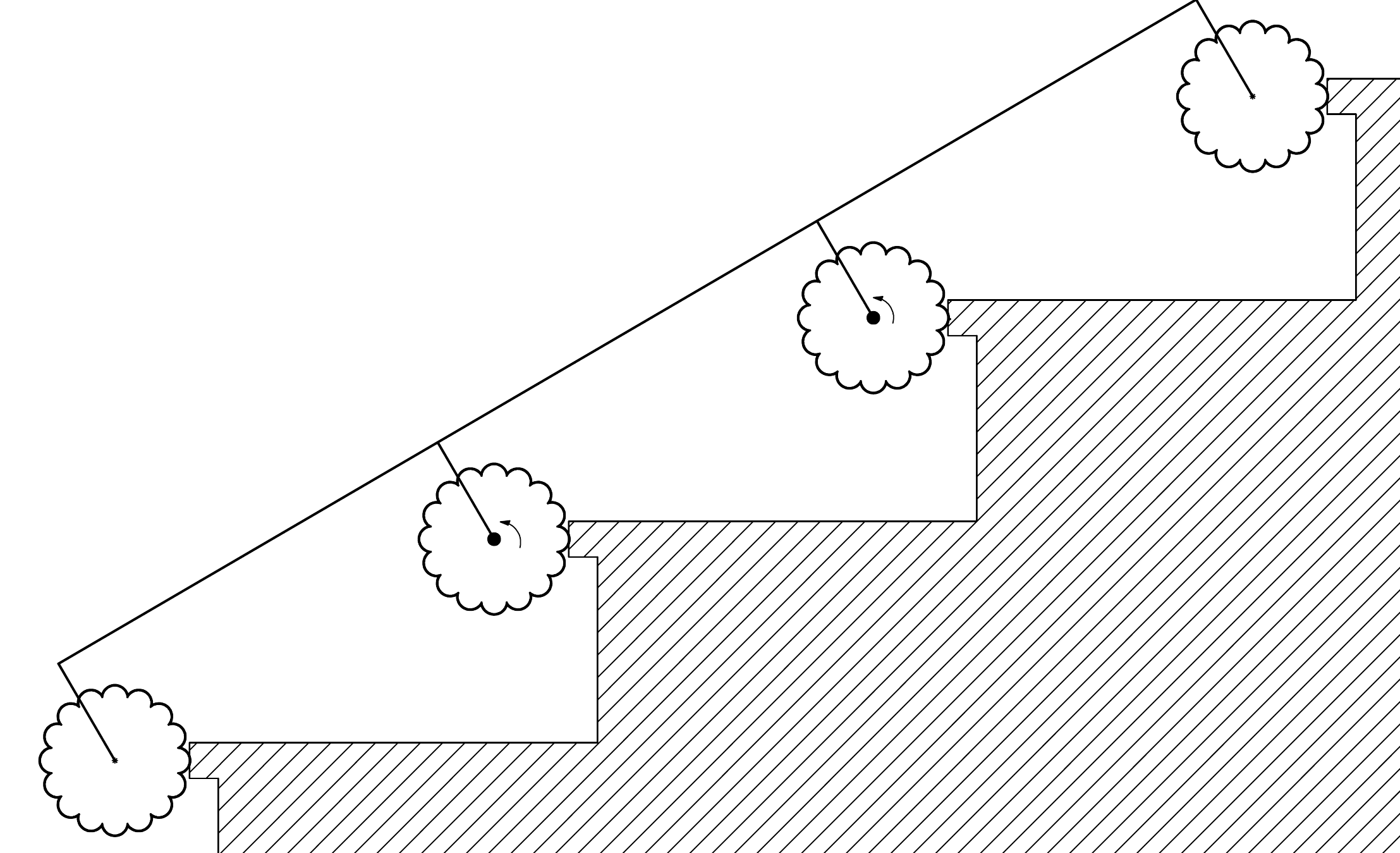} 
\caption{Concept for setting an upper bound on the module length}
\label{Lup}
\end{minipage}
\quad
\begin{minipage}[b]{0.45\linewidth}
\includegraphics[width=\linewidth ,trim = 50mm 80mm 70mm 30mm, clip]{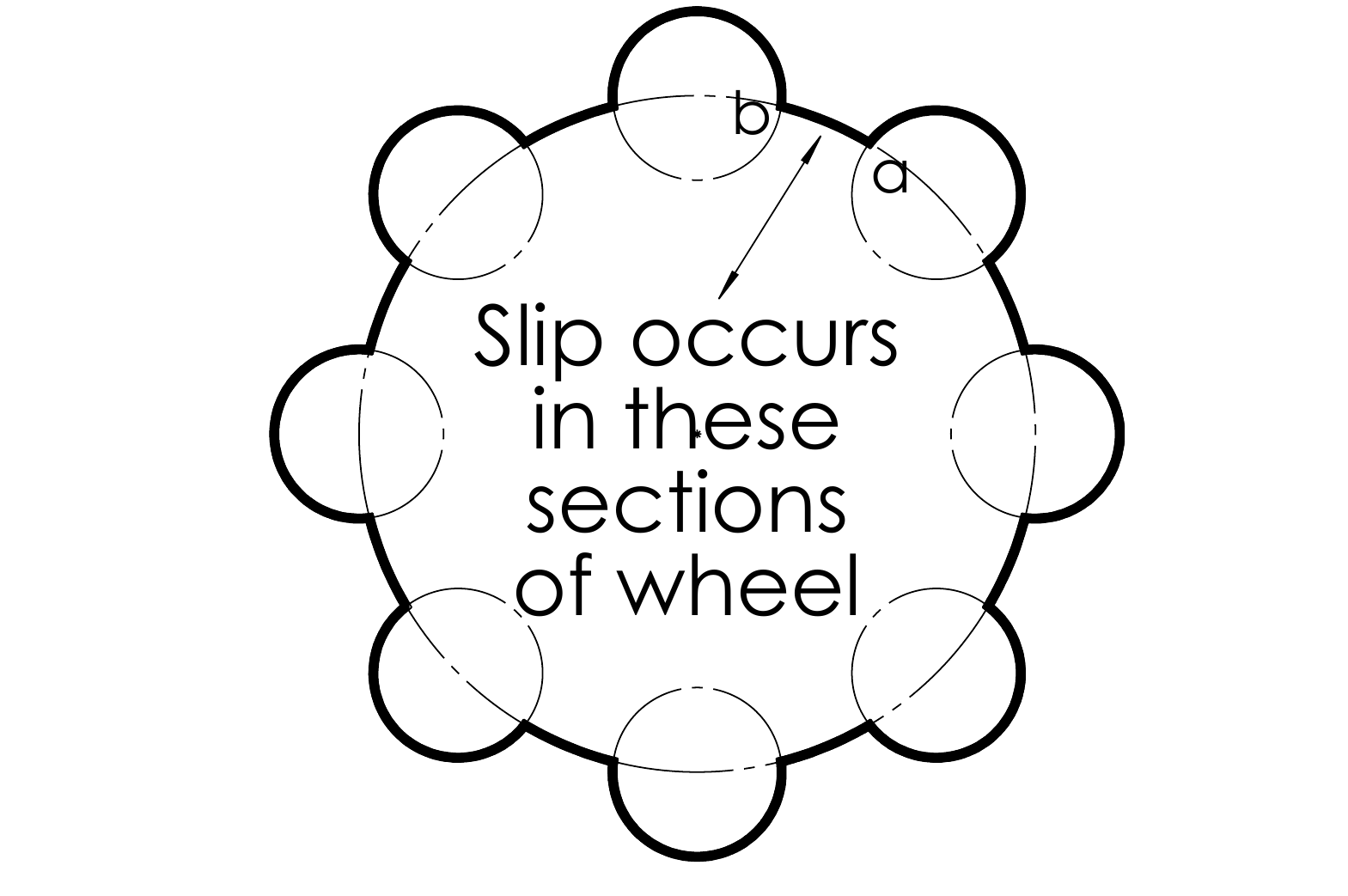}
\caption{Concept for setting a lower bound on the number of child circles}
\label{N_lb}
\end{minipage}
\vspace{-3.5mm}
\end{figure}
\begin{equation*}\label{Nlb}
\centering
n_{lb} = \lceil x \rceil \quad s.t. \quad x(2r_c) = 2\pi r_p 
\end{equation*}
Figure \ref{Nup1} depicts how as $n_c$ increases, the normal reaction generated moment gradually becomes clockwise from being counter-clockwise. The supremum for the number of child circles $n_{up}$ was chosen, such that the normal reaction obtained from the overhang generates a counter-clockwise moment about the axis of rotation of the module as can be seen in Figure \ref{Nup2}. This assures that even with minimal friction coefficient, the resultant contact force R generates a counter-clockwise moment. These conditions are contained in the following optimization:
\begin{figure}[t]
\centering
\resizebox{\linewidth}{!}{
  \begin{tabular}{ccc}
    \includegraphics[width=1\linewidth,trim = 220mm 210mm 250mm 120mm, clip]{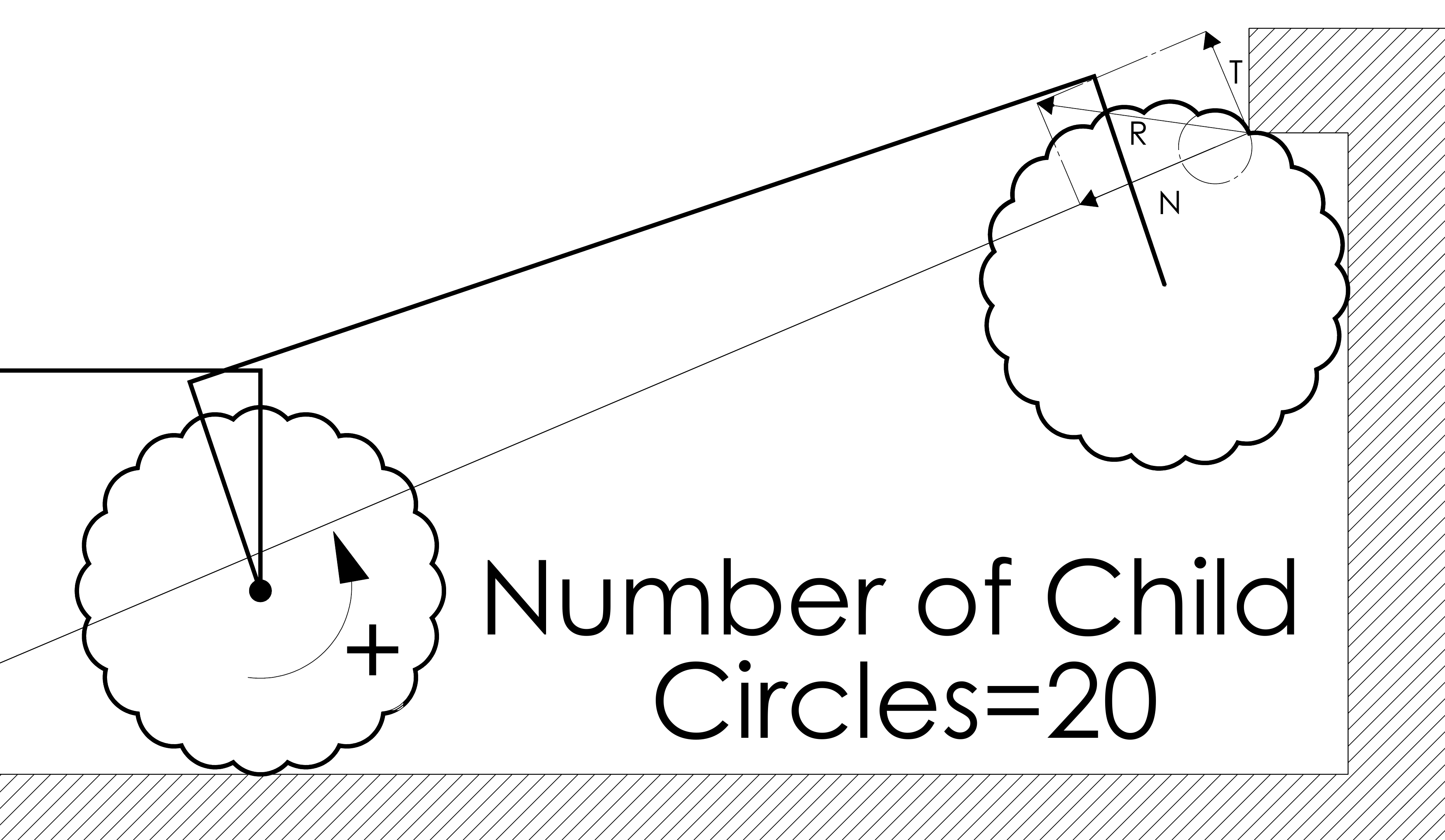} &
    \includegraphics[width=1\linewidth,trim = 220mm 210mm 250mm 120mm, clip]{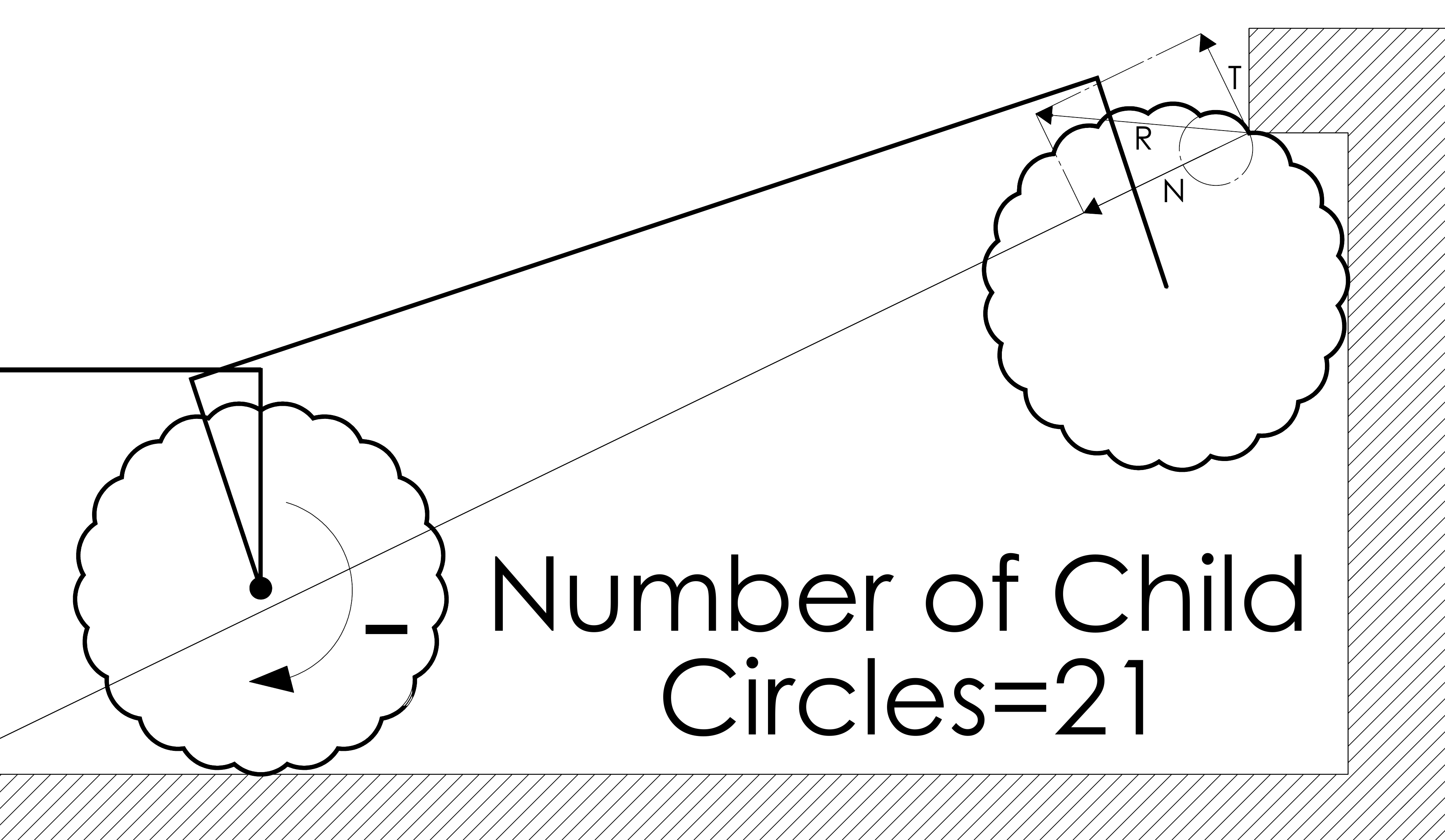} & \\
  \end{tabular}
  }
  \caption{Concept for setting an upper bound on the number of child circles}
  \label{Nup1}
  \vspace{-4mm}
\end{figure}
\begin{equation*}
     \begin{aligned}
     Maximize \ &  \ \ n_{up}\\
      Subject\ to\ &\\
       & (i)\ (\phi_1,\phi_2,\phi_3,\phi_4,\delta)  \in \mathbb{R} \\
       & (ii)\ (n_{up},p) \in \mathbb{Z} ^+ \\
       & (iii)\ r_p \cos (\phi_1) + r_c = r_p \cos ( \phi_1 + p \delta ) + \\
     & \qquad \qquad \qquad \qquad \qquad  r_c \cos (\phi_1 + p \delta - \phi_2 )\\  
      & (iv)\ r_p \sin (\phi_1 \pm \delta ) + r_c \leq r_p \cos (\phi_1) + r_c \\
     & (v)\ r_p \sin (\phi_1 + (p-1)\delta ) + r_c = \\
     & \qquad \quad r_p \sin ( \phi_1 + p \delta ) + r_c \sin (\phi_1 + p \delta - \phi_2 )\\
     & (vi)\ -\Big (\frac{\delta}{2}\Big ) < \phi_1 \leq \Big (\frac{\delta}{2}\Big ) \\
     & (vii)\ 0 \leq \phi_2 \leq \Big (\frac{\delta}{2} \Big ) + \sin^{-1} \Big ( \frac{r_p \sin (\frac{\delta}{2})}{r_c} \Big ) \\
     & (viii)\ p \leq \Big \lceil \Big (\frac{n_{up}}{4}\Big ) \Big \rceil \\
     & (ix)\  \phi_4 \leq \phi_3 \\
     \end{aligned}     
\end{equation*}
as shown in Figure \ref{Nup2}, $\phi_1$ is the angle made with the horizontal by the line joining the center of climbing wheel and the center of its child circle touching the face of the riser. $\phi_2$ is the angle between the line joining the lower tip of the overhang with the center of the child circle climbing the overhang and the line joining the center of the same child circle with the center of the parent circle. $\phi_3$ is the angle made by the line joining the center of the second wheel on the ground and the lower tip of the overhang with the horizontal. $\phi_4$ is the angle made by the normal reaction force obtained from the overhang with the horizontal. $\delta$ is the angle inscribed by the lines joining the center of the wheel and any two consecutive child circles. $p$ is the number of angles, $\delta$, inscribed by the lines joining the center of the parent circle and the centers of the child circle touching the face of the riser and the child circle touching the lower tip of the overhang. The constraints $(i)-(ii)$ represent the domain of the variables, $(iii)-(v)$ are the geometrical constraints, $(vi)-(viii)$ contain the range of the variables and the constraint $(ix)$ ensures that the normal reaction force generates the desirable counter-clockwise moment. Considering these bounds, the following range was obtained for the number of child circles:
\begin{equation} \label{Eq:N_bound}
\centering
16 \leq n_c \leq 20
\end{equation}
For the simplicity of the analysis, the control variables, clearance $c$ between the ground and the module link and the stiffness values of the springs at the joints were kept constant. The clearance $c$, for ensuring geometrical trafficability, was set, considering that the clearance in maximum when the wheel size is minimum and the module link length is maximum with an additional safety margin \cite{avi}.

The stiffness values for the springs at the joints, for all possible combinations of the control factors were calculated using the method used in \cite{harsha} and a mean value for all was considered for our analysis. Following are the values considered for the clearance and the spring stiffness at the joints: 
\begin{itemize}
  \item $c$ = $100$ $mm$
  \item $k^+_1$ = $74.47$ $Nmm$/$deg$ \quad $k^-_1$ = $67.56$ $Nmm$/$deg$
  \item $k^+_2$ = $48.89$ $Nmm$/$deg$ \quad $k^-_2$ = $57.61$ $Nmm$/$deg$
\end{itemize}
where, $k^+_i$ and $k^-_i$ are the stiffness values of the springs at the $i^{th}$ joint opposing the counter-clockwise and the clockwise moments respectively.

Three levels are set for each control parameter taking into consideration the equations (\ref{Eq:R_bound}) - (\ref{Eq:N_bound}). The levels, Level 1, Level 2 and Level 3, respectively set for the Module length are 240 $mm$, 260 $mm$ and 280 $mm$, that for Radius of the Parent Circle are 40 $mm$, 45 $mm$ and 50 $mm$, and finally for the  Number of Child Circles are	16, 18 and 20. 
\begin{table*}[t]
\vspace{1mm}
\centering
\caption{Orthogonal Array $L_9$($3^3$) experiments and their results, Grey Relational Coefficients and Grey Grades}
\label{OA_results}
\tiny
\resizebox{\linewidth}{!}{%
\renewcommand{\arraystretch}{1.2}
\begin{tabular}{ccccccccccccccc}
\hline
\begin{tabular}[c]{@{}c@{}}Run\\ No.\end{tabular} & A & B & C & \multicolumn{4}{c}{\begin{tabular}[c]{@{}c@{}}Power Required to\\ Climb 4 Steps (Watt)\end{tabular}} & \begin{tabular}[c]{@{}c@{}}Amplitude\\ (mm)\end{tabular} & Frequency & \multicolumn{3}{c}{Grey Relational Coefficients} & \begin{tabular}[c]{@{}c@{}}Grey\\ Relational Grade\end{tabular}  & Rank \\
                                                  &   &   &   & {\tiny $p^r_1$}                      & {\tiny $p^r_2$}                     & {\tiny $p^r_3$}                    & S/N Ratio&                                                    &      & Power   & Amplitude   & Frequency  &      &      \\ \hline
1                                                 & 1 & 1 & 1 & 4.02                    & 4.07                   & 5.19                   & -12.98                   & 0.76                                               & 16   & 0.33    & 0.45  & 0.86  & 0.59 & 5    \\
2                                                 & 1 & 2 & 2 & 3.12                    & 3.47                   & 3.35                   & -10.43                   & 0.68                                               & 18   & 0.67    & 0.55  & 0.79  & 0.57 & 7    \\
3                                                 & 1 & 3 & 3 & 3.49                    & 3.87                   & 3.79                   & -11.42                   & 0.61                                               & 20   & 0.48    & 0.65  & 0.72  & 0.49 & 8    \\
4                                                 & 2 & 1 & 2 & 2.96                    & 2.90                   & 3.52                   & -9.95                    & 0.60                                               & 18   & 0.83    & 0.67  & 0.71  & 0.66 & 3    \\
5                                                 & 2 & 2 & 3 & 2.99                    & 2.99                   & 3.08                   & -9.61                    & 0.55                                               & 20   & 1       & 0.79  & 0.64  & 0.70 & 2    \\
6                                                 & 2 & 3 & 1 & 3.57                    & 3.36                   & 3.61                   & -10.93                   & 0.96                                               & 16   & 0.56    & 0.33  & 1     & 0.63 & 4    \\
7                                                 & 3 & 1 & 3 & 4.28                    & 3.95                   & 3.70                   & -12.01                   & 0.49                                               & 20   & 0.41    & 1     & 0.55  & 0.58 & 6    \\
8                                                 & 3 & 2 & 1 & 3.33                    & 2.90                   & 3.44                   & -10.19                   & 0.86                                               & 16   & 0.74    & 0.38  & 0.94  & 0.70 & 1    \\
9                                                 & 3 & 3 & 2 & 3.96                    & 3.41                   & 4.15                   & -11.72                   & 0.75                                               & 18   & 0.44    & 0.46  & 0.86  & 0.47 & 9   \\ \hline
\end{tabular}
}
\vspace{2pt}
\caption*{{\small Legend: A: Levels of Module Length; B:Levels of Radius of Parent Circle; C:Levels of Number of Child Circles; $p^r_i$ is the power consumed by robot on the $i^{th}$ stair dimension }}
\vspace{-5mm}
\end{table*}
\section{Design of Experiments} \label{doe}
An all-inclusive experimentation of the control and noise factor levels under study would have required 81 experiments. Taguchi$'$s Orthogonal Array (OA) \cite{taguchi} provides a condensed set of experiments, which are balanced to ensure that all levels of factors are considered equally and can be evaluated independently of each other. Based on the number of parameters and their levels, the present experimentation was designed as per Taguchi$'$s $L_{9}$($3^3$) OA. It is quite time consuming, tedious and uneconomical to conduct physical experiments for all the combinations in the OA. For this reason, we simulated our experiments in $SolidWorks\ Motion\ Analysis$, which is a robust physics-based solver. Similar approach of using simulation methods in Taguchi's Method can be seen in the references \cite{rout, kim, lee}. The subsections ahead describe the experiments and the analysis performed on the experimental results. 
\subsection{Performance Metrics and their Data Collection}
Robustness can be achieved by identifying control factors which reduce variability in a product/process, by minimizing the effects of uncontrollable factors (noise factors). Noise factors cannot be controlled during product use, but can be controlled during experimentation. In a Taguchi designed experiment, noise factors are manipulated to force variability to occur and from the results, identify optimal control factor settings that make the process or product robust, or resistant to variation from the noise factors. Here, we vary the stair dimensions to force variability in the performance metrics. 

The performance values collected from each run in the OA are tabulated in Table \ref{OA_results}. The Signal to Noise ratio $(S/N)$, which reflects the ability of the system to perform well in the presence of noise factors, is considered as the metric for the power consumption attribute ($S/N$ ratio is not required for the Amplitude and Frequency attributes as they do not change with the varying stair dimensions). As we are interested in minimizing the power required, the ``smaller-the-better" formula for $S/N$ ratio given ahead is used
\begin{equation}\label{Eq:S/N}
\centering
S/N\ ratio = \eta = -10\log_{10} \left[\frac{1}{n} \sum\limits_{i=1}^{n} (p^r_i)^2 \right] \nonumber
\end{equation}
where, $\eta$ denotes the $S/N$ ratio calculated from the obtained $p^r_i$'$s$ which are the power consumption values from the $i^{th}$ run in the OA and $n$ is the number of times an experiment in the OA is repeated, i.e. the levels of noise factors, which is three here.
\vspace{-1mm}
\subsection{Multi-Attribute Decision Making Method}
Grey Relational Analysis (GRA) \cite{deng}, a multi-attribute decision making technique is used to obtain the optimum set of various input parameters for the best performance characteristics. GRA firstly translates performance of all alternatives into a comparability sequence, a normalized data sequence between 0 and 1. Following the normalization, Grey relational coefficients (GRC) are calculated to express the relationship between the ideal and normalized data sequence. A distinguishing coefficient is used to expand or compress the range of the GRC. We calculated the GRC considering a distinguishing coefficient of 0.5 and the grey relational grade for each combination of the control factors in the Orthogonal Array was obtained by equal weighing of their GRC. The experiments were then ranked according to their grey relational grades. Table \ref{OA_results} tabulates these results.

The influence of each control parameter on the Grey Relational Grade is presented in Table \ref{Influence}, providing the following optimal setting:
\begin{itemize}
  \item Module Length = $260$ $mm$,
  \item Radius of Parent Circle = $45$ $mm$ and
  \item Number of Child Circles = $16
$\end{itemize}
\begin{table}[t]
\centering
\caption{Influence of Control Parameters on Grey Relational Grade}
\label{Influence}
\resizebox{\linewidth}{!}{%
\renewcommand{\arraystretch}{1.5}
\begin{tabular}{lccc}
\hline
\begin{tabular}[c]{@{}l@{}}Control \\ Parameter\end{tabular} & \multicolumn{3}{c}{\begin{tabular}[c]{@{}c@{}}Average Grey\\   Relational Grade by factor level\end{tabular}} \\ \hline
                                                             & Level 1                              & Level 2                              & Level 3                         \\ \hline
Module Length                                      & 0.554162                             & 0.668827*                         & 0.587269                        \\
Radius of Parent Circle                                               & 0.615386                             & 0.664180*                         & 0.530693                        \\
Number of Child Circles                                      & 0.646039*                         & 0.570653                             & 0.593566                        \\ \hline
\multicolumn{4}{l}{Optimal levels are indicated by symbol *}                                                                                                                  
\end{tabular}
}
\vspace{-5mm}
\end{table}
\subsection{Explanation of the Grey Relational Analysis Results}
In all our simulation experiments, the angular velocity of all the wheels was kept constant. Therefore the power consumed by the robot is in proportion to the torque provided by the motors. As the wheel radius increases, the required driving torque also increases, thus increasing the power consumption. According to this argument, the minimum radius of the parent circle i.e. 40 mm should have been the optimal setting. But we also need to take into consideration the fact that as we go on decreasing the radius of the parent circle, it becomes more difficult for the contact force on the wheel to provide a counter-clockwise moment about the axis of rotation of the module while overcoming the overhang. Hence, neither the least and nor the largest, but the nominal level of the parameter performed optimally. Same was the case with the module length. Increasing the length of the module results in configurations which enable more number of wheels to be in contact with the step tread providing a better pushing and pulling force against the step risers on the wheels climbing the risers. But, at the same time, increase in the module length also results in an increased power consumption. Thus a nominal value for the module length resulted as the optimal setting. As the number of child circles decreases, the normal force obtained from the overhang gets directed farther away from the axis of rotation of the module thus making it easier for the module to overcome the overhang. Hence, the least value for the number of child circles resulted in the optimal performance.
\subsection{Analysis of Variance}
\begin{table}[]
\centering
\caption{Analysis of Variance}
\label{ANOVA}
\resizebox{\linewidth}{!}{%
\renewcommand{\arraystretch}{1.3}
\begin{tabular}{lcccc}
\hline
                                                                   & \begin{tabular}[c]{@{}c@{}}Module\\ Length\end{tabular} & \begin{tabular}[c]{@{}c@{}}Radius of\\ Parent Circle\end{tabular} & \begin{tabular}[c]{@{}c@{}}Number of\\ Child Circles\end{tabular} & Error    \\ \hline
\begin{tabular}[c]{@{}l@{}}Sum of\\ Squares\end{tabular}           & 0.020896                                                & 0.027373                                                          & 0.008962                                                          & 0.001618 \\ \hline
F-test                                                             & 12.91703                                                & 16.9208                                                           & 5.539721                                                          &          \\ \hline
\begin{tabular}[c]{@{}l@{}}Percentage \\ Contribution\end{tabular} & 35.50825                                                & 46.5144                                                           & 15.2284                                                           & 2.748948 \\ \hline
\end{tabular}
}
\vspace{-5mm}
\end{table}
The variability in the response variable, amongst different factors is decomposed using  a statistical tool, Analysis of Variance (ANOVA) \cite{anova}. The F-test \cite{anova} was used to study the statistically significant control factors in our analysis. The results of the test are presented in Table \ref{ANOVA}, and indicate that the Radius of the Parent Circle and the Length of the Module play a statistically significant role in the performance measure of the vehicle.
\subsection{Prediction of Grey Relational Grade}
The Grey Relational Grade can be predicted for the optimal setting using the following equation
\begin{equation}\label{Eq:Predict}
\centering
\gamma_p = \gamma_m + \sum\limits_{i=1}^{n} (\gamma_m -\gamma_i) 
\end{equation}
where, $\gamma_p$ is the predicted Grey Relational Grade for the optimal setting, $\gamma_m$ is the mean Grey Relational Grade, n is the number of statistically significant factors affecting the performance and $\gamma_i$ is the maximum average Grey Relational Grade of a level for the $i^{th}$ statistically significant factor. The application of this to our case is presented in the immediate subsection.
\subsection{Confirmation in Simulation}
\begin{table}[]
\centering
\caption{Optimal Setting Confirmation}
\label{opt_conf}
\resizebox{0.9\linewidth}{!}{%
\renewcommand{\arraystretch}{1.2}
\begin{tabular}{lccc}
\hline
                                                                       & \begin{tabular}[c]{@{}c@{}}Best OA \\ Expt.\end{tabular} & \multicolumn{2}{c}{Optimal Setting} \\ \hline
                                                                       &                                                          & Prediction       & Simulation       \\ 
Level                                                                  & $A_3B_2C_1$                                                   & $A_2B_2C_1$ & $A_2B_2C_1$         \\ \hline
\begin{tabular}[c]{@{}l@{}}Power Consumption \\ S/N Ratio\end{tabular} & -10.1992                                                 &                  & -9.4244          \\ 
Amplitude                                                              & 0.864662                                                 &                  & 0.864662         \\ 
Frequency                                                              & 16                                                       &                  & 16               \\ 
\begin{tabular}[c]{@{}l@{}}Grey Relational\\  Grade\end{tabular}       & 0.709478                                                 & 0.729588         & 0.795383         \\ \hline
\end{tabular}
}
\vspace{-5mm}
\end{table}
A simulation experiment was performed at the optimal setting of the control factors and its calculated Grey Relational Grade was compared with the Predicted Grey Relational Grade, the results of which are tabulated in the Table \ref{opt_conf}. Its closeness to the predicted grade, successfully validates our analysis. 
\section{Confirmation in Experiment}\label{exp}

A robot prototype with the set of optimal design parameters obtained from the proposed framework was manufactured. The material of choice for the chassis was aluminium to cater to low weight and the wheels were 3D printed. The wheels were rubber padded for improved traction while climbing. Compliance was added by attaching a set of torsion springs at each module joint. The motors were chosen such that they provide enough torque to achieve the desired moments for the modules to fold with respect to each other and to carry the robot's body upward to climb upstairs.
\begin{figure}[h]
\centering
\includegraphics[width=0.45\linewidth]{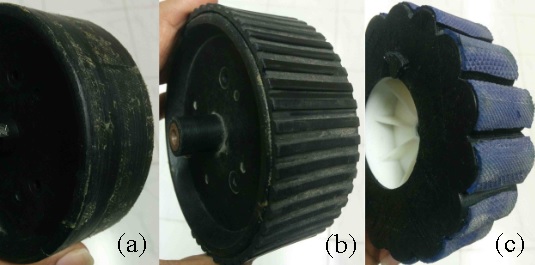} 
\caption{The different wheels experimented with to check climbing ability}
\label{wheels}
\vspace{-4mm}
\end{figure}
\begin{figure}[h]
\centering
\includegraphics[width=0.8\linewidth]{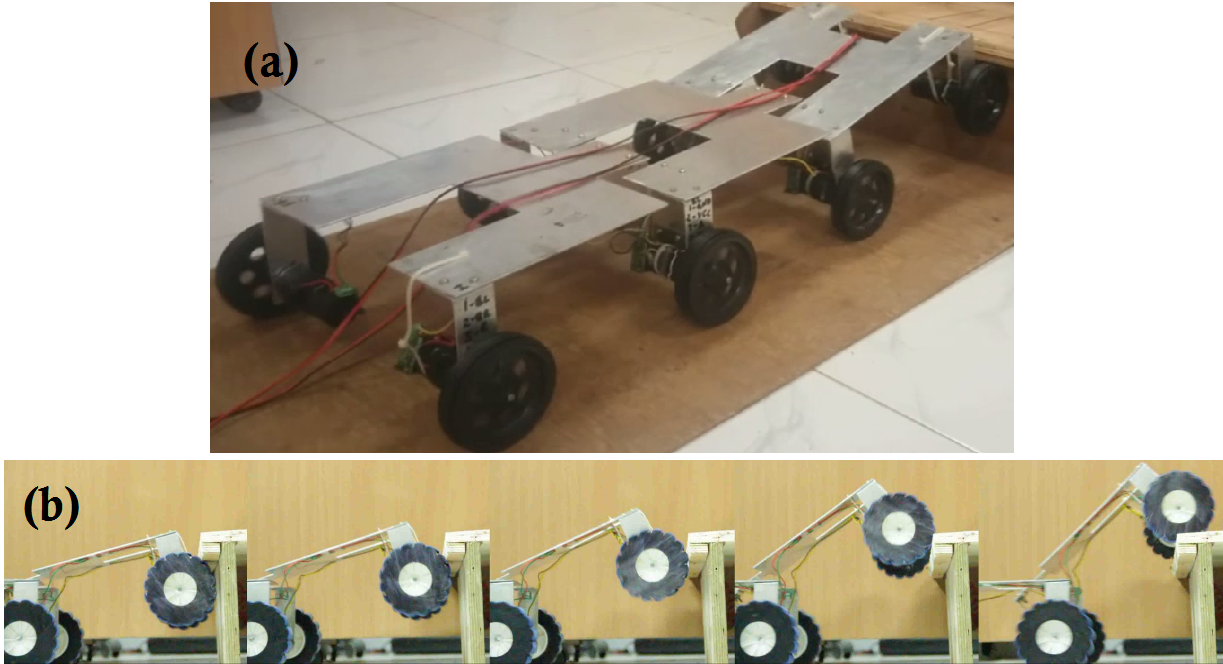} 
\caption{(a) The robot prototype with circular wheels getting jammed under the overhang (b) The modified wheel tackling an overhang owing to the circular arcs on the periphery}
\label{jammed}
\vspace{-6mm}
\end{figure} 

\begin{figure}[htb]
\centering
\includegraphics[width=\linewidth]{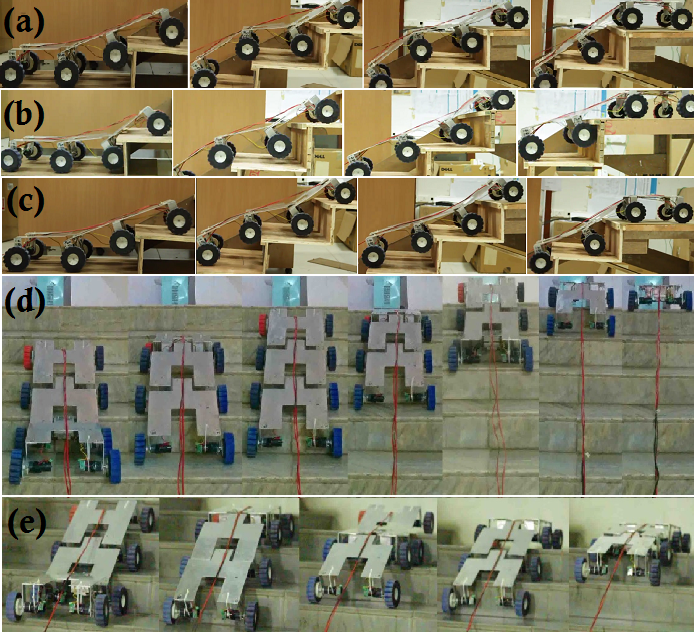} 
\caption{The robot prototype with optimal set of design parameters successfully climbing stairs of dimensions specified in the first,second and third level of noise factors ((a),(b) and (c)), and on staircases in urban setting ((c) and (d))}
\label{Simfig2}
\vspace{-7mm}
\end{figure}

The robot equipped with a normal circular wheel with smooth rubber padding (Figure \ref{wheels} (a)) failed to climb the stairs with overhang, as shown in Figure \ref{jammed} (a). According to the reasoning provided in Section \ref{conceptdesign}, the robot fitted with wheels with low child circle radius (Figure \ref{wheels} (b)) also kept on slipping at the overhang, as the rubber gets deformed and fails to provide the desired counter-clockwise moment. Figure \ref{jammed} (b) shows the modified wheel (Figure \ref{wheels} (c)) overcoming an overhang, thus validating our concept design. The robot, without getting jammed under the overhang or getting stuck in any configuration (Figure \ref{Simfig2}), successfully climbed the stair dimensions considered in our analysis, as well as, staircases in urban setting. We have even developed a standalone version of this prototype equipped with batteries which can independently climb stairs. It's performance is captured in our video submission.\vspace{-1mm}
\section{Conclusions and Future Work} \label{future}
In this paper we report the challenge of overcoming overhang for Class I robots i.e. robots with wheel diameter lesser than the step riser dimension and tackle this problem for our robot by bringing a modification in the wheel design. The rigorous analysis and the kind of formulation of wheel design as an optimization problem has not been seen before. We also dress the robot with robust design parameters enabling the robot to climb stairs of varying dimensions and make its performance least susceptible to this variation. It was concluded that it is practicable to use
a statistical method to solve the optimality of the robot given the complex design and the subsequent analysis that would have necessitated, had it been posed as a non-linear optimization problem. 

The varying stair dimensions considered in our analysis accord with the IBC thus providing a high impact work for urban search and rescue operations. Simulation and experimental results corroborate the analysis performed as the robot was able to scale numerous overhang stairs with varying step dimensions.

We would focus our future work on developing actively controlled joints for a modular robot. Furthermore, we wish to add joints to enable the robot to turn and consequently climb Circular Stairs and extremely non-trivial terrains.

\bibliographystyle{unsrt}
\bibliography{overhang}

\end{document}